\documentclass[twoside,leqno,twocolumn]{article}

\usepackage[letterpaper]{geometry}
\usepackage{verbatim}
\usepackage{ltexpprt}
\usepackage{hyperref}
\usepackage{cite}
\usepackage{amsmath,amssymb,amsfonts}
\usepackage{graphicx}
\usepackage{textcomp}
\usepackage{xcolor}
\usepackage{xspace}
\usepackage{algorithm}
\usepackage{algpseudocode}

\algrenewcommand\alglinenumber[1]{#1}

\usepackage{amsthm}
\usepackage{flushend}

\usepackage{graphicx}
\usepackage{multirow}
\usepackage{tabularx}
\usepackage{subcaption}
\usepackage{booktabs}

\newtheoremstyle{def_style}
  {}          
  {}          
  {}          
  {}          
  {\bfseries} 
  {.}         
  {.5em}      
  {}          

\theoremstyle{def_style}

\theoremstyle{def_style}

\theoremstyle{def_style}

\theoremstyle{def_style}

\newcommand{\ours}{\texttt{pFedKnow}\xspace}

\begin{document}
\title{Knowledge-Enhanced Semi-Supervised Federated Learning for Aggregating Heterogeneous Lightweight Clients in IoT}
\author{Jiaqi Wang\thanks{Pennsylvania State University. Email: jqwang@psu.edu. The frist two authors have equal contributions to this work.}
\and Shenglai Zeng\thanks{University of Electronic Science and Technology of China. Email: shenglaizeng@gmail.com}
\and Zewei Long \thanks{University of Illinois Urbana-Champaign. Email: zeweil2@illinois.edu}
\and Yaqing Wang\thanks{Purdue University. Email: wang5075@purdue.edu}
\and Houping Xiao\thanks{Georgia State University. Email: hxiao@gsu.edu}
\and Fenglong Ma\thanks{Pennsylvania State University. Email: fenglong@psu.edu}
}
\date{}
\maketitle

\fancyfoot[R]{\scriptsize{Copyright \textcopyright\ 2023 by SIAM\\
Unauthorized reproduction of this article is prohibited}}

\begin{abstract} 
Federated learning (FL) enables multiple clients to train models collaboratively without sharing local data, which has achieved promising results in different areas, including the Internet of Things (IoT). 
However, end IoT devices  do not have abilities to automatically  annotate their collected data, which leads to the label shortage issue at the client side. To collaboratively train an FL model, we can only use a small number of labeled data stored on the server. This is a new yet practical scenario in federated learning, i.e., labels-at-server semi-supervised federated learning (SemiFL). 
Although several SemiFL approaches have been proposed recently, none of them can focus on the personalization issue in their model design.  IoT environments make SemiFL more challenging, as we need to take device computational constraints and communication cost into consideration simultaneously.
To tackle these new challenges together, we propose a novel SemiFL framework named \ours. \ours generates lightweight personalized client models via neural network pruning techniques to reduce communication cost. Moreover, it incorporates pretrained large models as prior knowledge to guide the aggregation of personalized client models and further enhance the framework performance. Experiment results on both image and text datasets show that the proposed \ours outperforms state-of-the-art baselines as well as reducing considerable communication cost. The source code of the proposed \ours is available at \url{https://github.com/JackqqWang/pfedknow/tree/master}.

\end{abstract}

\begin{figure*}[!t]
    \centering
    \includegraphics[width=0.85\textwidth]{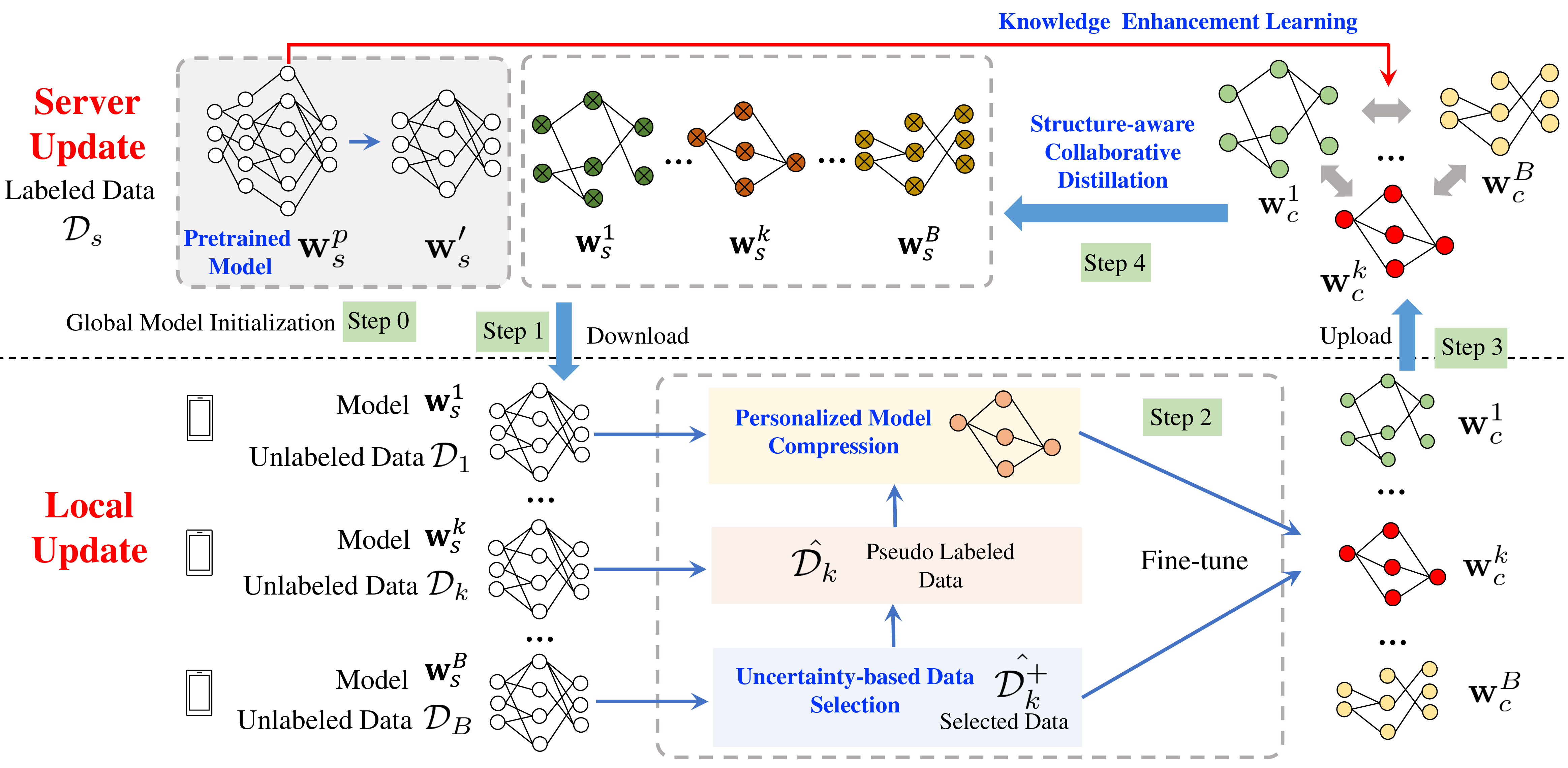}
    \caption{Overview of the proposed \ours model, which includes initialization, local personalized model update, and server update using structure-aware collaborative distillation and knowledge enhancement learning.}
    \label{fig:framework}
\end{figure*} 

\section{Introduction} 


Federated learning (FL) aims to protect user data privacy by training a shared global model under the coordination of a central server while keeping the user data decentralized on local clients~\cite{mcmahan2017communication}. 
This new learning paradigm has been widely applied in the Internet of Things (IoT) area~\cite{yuan2020federated, samarakoon2018federated,savazzi2020federated,khan2021federated}. 
However, most of the current studies in IoT assume that the data stored in local devices are fully labeled. 
This assumption is impractical since users usually do not take any incentives or have the expertise to annotate the generated data. 
Thus, training with unlabeled client/user data seems to be more practical for real-world IoT applications. 
In order to provide better service, attract more users, and further increase profits, companies may be willing to annotate a small set of data and put them on a server to help train a more accurate FL model.
This leads to a new scenario in FL, which is \emph{labels-at-server semi-supervised federated learning} (SemiFL).

Recently, a few SemiFL approaches have been proposed, such as FedMatch~\cite{jeong2020federated}, SSFL ~\cite{zhang2020improving}, FedMix ~\cite{zhang2021semi}, FedSEAL ~\cite{bian2021fedseal}, SemiFL ~\cite{diao2022semifl}, and FedTriNet~\cite{che2021fedtrinet}. 
These models primarily focus on developing a good global model and serving one global model to all clients. However, a general global model may not be sufficient to characterize the uniqueness of each IoT user since IoT devices may store heterogeneous data. Thus, model customization has become a rigid need for IoT applications.
Moreover, existing methods are not developed for IoT applications and do not take into account the constraints of IoT devices such as limited computational resources and constrained network bandwidth.
Thus, it is an urgent need to develop a new semi-supervised FL paradigm in IoT.

However, we will face several challenges to develop such a method. 
First, there is no labeled data in the labels-at-server setting, which makes it impossible to directly apply existing personalized FL techniques such as pFedMe~\cite{t2020personalized}, since they all need labeled client data.  
Second, to save computational resources and reduce communication cost, IoT applications usually use models with less parameters, i.e., \emph{lightweight} models. Those models are hard to maintain competitive performance with the large ones. Therefore, the new challenges are how to guarantee the performance of the new SemiFL model and achieve the personalization of client models simultaneously.

To address the aforementioned challenges, in this paper, we propose a novel SemiFL framework named {\ours} for the labels-at-server scenario in IoT. 
In particular, we propose to introduce the advanced \emph{network structure compression} techniques~\cite{liu2017learning, xu2021rethinking} into our designed \ours for addressing the challenges we talked about. The motivation behind this design is that the trained network parameters of each client model are data-dependent, which provides a new way to achieve personalization by customizing different model structures for different clients.
The benefits are two folded: 
(1) Compressing network structures mainly depends on the unique characteristics of data stored at each client, which enables to generate personalized client models. Since the data stored at each client are different, the network structure of the compressed models at each client will also be different, even using the same initialized lightweight model. In such a way, for each client, we can obtain a personalized model. (2) Compressing lightweight models can further reduce communication cost between clients and the server. 

Network structure compression is a double-edged sword, which will significantly increase the difficulty of aggregating heterogeneous structured client models at the server side. 
To solve this issue, we propose a novel structure-aware collaborative distillation mechanism to produce aggregated personalized models with the same structures as the corresponding client models. We propose to equip the server with a powerful \emph{pretrained large model} as prior knowledge to further enhance the performance of our designed \ours. 
To be specific, the pretrained model is used not only to initialize the lightweight global model using knowledge distillation techniques but also to participate in and guide the aggregation of personalized client models on the server.


Figure~\ref{fig:framework} shows the overview of the proposed framework {\ours}, which consists of two modules including both the local update and server update with five steps. At step \textcircled{0}, we first initialize a lightweight global model $\mathbf{w}_s^{\prime}$ on the server by training on the labeled data $\mathcal{D}_s$ with the help of the pretrained model $\mathbf{w}_s^p$. This lightweight global model will be distributed to $B$ selected clients (i.e., step \textcircled{1}), where $B \ll K$ and $K$ is the number of total clients. 
Each client $k$ will update a local model $\mathbf{w}^k_c$ using the unlabeled data $\mathcal{D}_k$ and the downloaded model $\mathbf{w}_s^k$ from the server (i.e., step \textcircled{2}). Note that in the first communication round, $\mathbf{w}_s^k = \mathbf{w}_s^{\prime}$.
To learn a lightweight personalized local model on each client $k$, we propose a novel personalization approach for all local models and thus output $\{\mathbf{w}_c^1, \cdots, \mathbf{w}_c^B\}$ with different network architectures. These personalized models will be uploaded to the server for aggregation (i.e., step \textcircled{3}). Different from all the existing work such as FedAvg~\cite{mcmahan2017communication}, we propose a structure-aware collaborative distillation and enhancement learning approach to obtain a personalized local model on the server with the help of other uploaded client models and the original large model (i.e., step \textcircled{4}). These distilled models will be distributed to the selected clients in the following rounds. {\ours} will iteratively run from step \textcircled{1} to step \textcircled{4} until it converges. 

To sum up, the contributions of this work are listed as follows: 
\begin{itemize}
    \item To the best of our knowledge, we are the first work to distill lightweight models to warm up and further customize compressed local models with different structures using network pruning techniques in FL, which further solves the challenges of the limited local device computational capacity and restricted network bands in IoT.
    \item We propose a new aggregation approach with the combination of network structure-aware collaborative distillation  and large-model knowledge enhancement learning, which can obtain a personalized model with the help of other models even with different structures and extract general knowledge from pretrained large models.
    \item We conduct extensive experiments on both image and text datasets to show the effectiveness and efficiency of the proposed {\ours} framework compared with state-of-the-art baselines. 
\end{itemize}

\section{Task \& Notations}
In this paper, we focus on the realistic labels-at-server semi-supervised learning scenario. Under this setting, the server stores a small set of labeled data denoted as $\mathcal{D}_s = \{(x_i^s, y_i^s)\}_{i= 1}^{N_s}$, where $x_i^s$ is the data sample, $y_i^s$ is the corresponding label, and $N_s$ is the number of labeled data. Assume that there are $K$ clients, and each client $k$ only obtains a large number of unlabeled data denoted as $\mathcal{D}_k = \{x_i^k\}_{i= 1}^{N_k}$, where $N_k$ is the number of unlabeled data stored in the $k$-th client and $N_k \gg N_s$. 
The \textbf{goal} of this paper is to jointly train personalized client models $[\mathbf{w}_c^1, \cdots, \mathbf{w}_c^K]$ using the pretrained model $\mathbf{w}_s^p$ and labeled data $\mathcal{D}_s$ on the server as well as the unlabeled data $[\mathcal{D}_1, \cdots, \mathcal{D}_K]$ stored in all clients. 
We summarize the key notations used in the following sections in Table~\ref{tab:notation}.

\begin{table}[!tbp]
\caption{Key Notations.}
\resizebox{0.48\textwidth}{!}{
\begin{tabular}{cl}
\toprule
Symbol  & Definition and description \\
          \midrule
$\mathcal{D}_s$ & the set of labeled data on the server   \\
$\mathcal{D}_k$ & the set of unlabeled data on the $k$-th client\\
$\hat{\mathcal{D}}_k^{+}$ & the set of high-quality pseudo labeled data at client $k$\\
$K$ & the number of clients\\
$B$ & the number of active/selected clients\\

$\mathbf{w}_s^p$ & the pretrained model \\
$\mathbf{w}_s^\prime$ & the initialized global model \\
$\mathbf{w}_s^k$ & distilled model parameters for the $k$-th client on server\\
$\mathbf{w}_c^k$ & the $k$-th client's local model parameters\\
$\boldsymbol{\alpha}_i^k$ & logit outputted by $\mathbf{w}_s^k$ for the unlabeled data $x_i^k$ \\
$u_i^k$ & uncertainty score for the unlabeled data $x_i^k$\\
\bottomrule
\end{tabular}}
\label{tab:notation}
\end{table}

\section{Methodology}

\subsection{Initialization}
Since there is no labeled data stored in each client, we need to start the model learning from the server side. To be specific, we first initialize a shared global model $\mathbf{w}_s^\prime$ on the server using the labeled data $\mathcal{D}_s$, which is then distributed to each client. 
A naive solution is to directly train $\mathbf{w}_s^\prime$ only using $\mathcal{D}_s$. 
However, the small number of labeled data may lead to the underfitting issue. To solve this problem, we propose to equip a large pretrained model $\mathbf{w}_s^p$ on the server, which helps to initialize a better lightweight model $\mathbf{w}_s^\prime$ via knowledge distillation techniques as follows:
\begin{equation}\label{eq:init_kd}
\mathcal{L}_{ini} = {\rm CE}\left(f(\mathbf{x}^s; \mathbf{w}_s^\prime, \mathbf{y}^s\right) + {\rm KL}(\boldsymbol{\alpha}_s^p, \boldsymbol{\alpha}_s^\prime),
\end{equation}
where ${\rm CE}$ is the cross entropy loss, $f(\cdot;\cdot)$ represents the lightweight neural network, $\mathbf{x}^s$ represents the data features, and $\mathbf{y}^s$ is the label vector. $\rm KL$ denotes the Kullback-Leibler divergence. $\boldsymbol{\alpha}_s^p$ and $\boldsymbol{\alpha}_s^\prime$ represent the logits outputted from the pretrained model $\mathbf{w}_s^p$ and the lightweight initial model $\mathbf{w}_s^\prime$, respectively. 
\ours is a general framework and can deal with different types of data. Thus, we will choose different pretrained models and lightweight client models in our experiments.

The learned lightweight model $\mathbf{w}_s^{\prime}$\footnote{In this paper, model parameter set $\mathbf{w}$ is also used to represent the model architecture.} will be distributed to all local clients for the local model learning in the first training round.  
From the second round, each local model will receive an updated personalized model $\mathbf{w}_s^k$, which will be discussed in Section~\ref{sec:server_side}. Without the loss of generality, we will use $\mathbf{w}_s^k$ as the distributed local model from the server for each client $k$ in the following sections. 

\subsection{Model Personalization (Local Update)}
Following the training procedure of general federated learning models, a small subset of clients with size $B$ will be randomly selected to conduct local model learning. Assume that the $k$-th client is selected, and the unlabeled data $\mathcal{D}_k$ will be used for its local model updates.

\subsubsection{Personalized Model Compression}\label{sec:PMC}
Since there are only unlabeled data stored in each client, traditional supervised learning techniques~\cite{mcmahan2017communication,
smith2017federated, finn2017model, li2020federated,li2019fedmd} will not work. Thus, it is essential to design a new learning paradigm to incorporate unlabeled data into model learning. An intuitive way is to generate pseudo-labels for unlabeled data based on the downloaded model $\mathbf{w}_s^k$ from the server. These pseudo-labeled data will further be used for local model training. Let $\hat{y}_i^k$ denote the pseudo label predicted by $\mathbf{w}_s^k$ for each unlabeled data $x_i^k$, i.e., $\hat{y}_i^k = f(x_i^k; \mathbf{w}_s^k)$, and $\hat{\mathcal{D}}_k = \{(x_i^k, \hat{y}_i^k)\}_{i=1}^{N_k}$ be the set of pseudo labeled data.

To reduce the communication cost between clients and the server, an easy way is to reduce the number of parameters, i.e., learning lightweight local models. Model compression is a state-of-the-art approach to learning a performance-comparable lightweight model, which is commonly used in computer vision~\cite{wu2017squeezedet,yu2017compressing,gong2014compressing,frankle2018lottery} and natural language processing~\cite{lobacheva2017bayesian, jiao2019tinybert, xu2021rethinking, zafrir2019q8bert}. In other words, the lightweight models should have \emph{similar predictive ability} to the original larger model. This motivates us to learn a lightweight $\mathbf{w}_c^k$ even using the pseudo labeled data $\hat{\mathcal{D}}_k$. 

Thus, on each client $k$, we directly apply existing \textbf{network pruning} techniques~\cite{liu2017learning,xu2021rethinking} to learn a lightweight model $\mathbf{w}_c^k$ based on the server downloaded model $\mathbf{w}_s^k$ and the pseudo labeled data $\hat{\mathcal{D}}_k$, i.e.,
\begin{equation}
    \mathbf{w}_c^k = g(\hat{\mathcal{D}}_k; \mathbf{w}_s^k, \gamma),
\end{equation}
where $g(\cdot; \cdot)$ is the compression function and $\gamma$ is a hyperparameter to control the compression ratio. For image data, we apply the network slimming approach~\cite{liu2017learning} and the weight pruning part in SparseBERT~\cite{xu2021rethinking} for text data. Note that the lightweight model $\mathbf{w}_c^k$ holds the key architectures of $\mathbf{w}_s^k$ but with fewer parameters.

Since the pseudo labeled data $\hat{\mathcal{D}}_k$ $(k = 1, \cdots, B)$ on different local clients are different, even using the same model compression technique $g(\cdot; \cdot)$, we cannot guarantee that the compressed lightweight models $\{\mathbf{w}_c^1, \cdots, \mathbf{w}_c^B\}$ have the same network architecture. Such unique characteristic benefits us to conduct \emph{personalized semi-supervised federated learning} on each client.

Directly uploading $\mathbf{w}_c^k$ to the server will definitely reduce the communication cost. However, the current local model $\mathbf{w}_c^k$ only mimics the prediction behaviors of $\mathbf{w}_s^k$ on the pseudo labeled data $\hat{\mathcal{D}}_k$, which does not really update the model parameters with the help of local data. Unfortunately, we cannot use all of the data in $\mathbf{w}_s^k$ for updating $\mathbf{w}_c^k$ since the pseudo label quality of $\hat{y}_i^k$ may vary a lot. To address this issue, we propose to use uncertainty scores to filter out pseudo labels with low quality. The smaller the uncertainty score, the higher the quality of the pseudo labels.

\subsubsection{Uncertainty-based Data Selection}
Let $\boldsymbol{\alpha}_i^k \in \mathbb{R}^{|\mathcal{C}|}$ denotes the learned logit using $\mathbf{w}_s^k$ on the unlabeled data $x_i^k$, where $|\mathcal{C}|$ is the number of label category. For different types of data, $\mathbf{w}_s^k$ has different network architectures. Thus, we will use different ways to generate uncertainty scores, but they are all based on the learned logit $\boldsymbol{\alpha}_i^k$.

For image data, we utilize the approach proposed in the previous work~\cite{sensoy2018evidential} to quantify the uncertainty score as follows:
\begin{equation}\label{eq:image_uncertainty}
    u_i^k = \frac{|\mathcal{C}|}{\sum_{c=1}^{|\mathcal{C}|}({\rm ReLU}(\alpha^k_{i, c}) + 1)}.
\end{equation}
For text data, we directly use the probability distribution $[p^k_{i, 1}, \cdots, p^k_{i, |\mathcal{C}|}]$ learned using $\boldsymbol{\alpha}_i^k$ over a softmax layer for each unlabeled data $x_i^k$. Then, the uncertainty score is modeled as follows:
\begin{equation}\label{eq:text_uncertainty}
    u_i^k = 1 - \max\{p^k_{i, 1}, \cdots, p^k_{i, |\mathcal{C}|}\}.
\end{equation}
If $u_i^k \leq \delta$, then $(x_i^k, \hat{y}_i^k)$ will be selected for local model update, where $\delta$ is a threshold hyperparameter. To avoid ambiguity, we use $(x_i^k+, \hat{y}_i^k+)$ to represent the selected data. This uncertainty-based data selection approach allows us to obtain a high-quality pseudo label-based dataset $\hat{\mathcal{D}}_k^{+}  = \{(x_i^k+, \hat{y}_i^k+)\}_{i=1}^{N_k^{+}} \subseteq \hat{\mathcal{D}}_k$ on each client $k$, where $N_k^{+}$ $(\leq N_k)$ is the number of selected data. 

\subsubsection{Personalized Model Update}
Based on the selected high-quality pseudo data $\hat{\mathcal{D}}_k^{+}$, we can fine-tune the lightweight model $\mathbf{w}_c^k$ using the following loss function:
\begin{equation}\label{eq: pmu}
    \mathcal{L}_k = {\rm CE}\left(f_k\left(\mathbf{x}^k+; \mathbf{w}_c^k\right), \hat{\mathbf{y}}^k+\right),
\end{equation}
where $\mathbf{x}^k+$ is the data features and $\hat{\mathbf{y}}^k+$ is the corresponding high-quality pseudo label vector. 
Finally, these fine-tuned lightweight model parameters $\{\mathbf{w}_c^k, \cdots, \mathbf{w}_c^B\}$ will be uploaded to the server. Next, we will introduce how to conduct model fusion in the next subsection.

\subsection{Collaborative Distillation (Server Update)}\label{sec:server_side}

After the server receives local uploaded models $\{\mathbf{w}_c^1, \cdots, \mathbf{w}_c^B\}$, if following existing federated learning approaches such as FedAvg~\cite{mcmahan2017communication}, we will aggregate $\{\mathbf{w}_c^1, \cdots, \mathbf{w}_c^B\}$ to generate a shared global model and then distribute it to each local again.
However, as mentioned in Section~\ref{sec:PMC}, the network architectures of lightweight models $\{\mathbf{w}_c^1, \cdots, \mathbf{w}_c^B\}$ may be significantly different from each other, which leads to the failure of existing model aggregation approaches. Thus, we need to design a new learning paradigm to conduct heterogeneous model fusion on the server.

However, it is non-trivial to design such a new model since it will satisfy the following requirements: First, for personalized learning, we need to maintain the key characteristics of the current model $\mathbf{w}_c^k$ as well as taking other uploaded model parameters $\{\mathbf{w}_c^j\}_{j=1}^{B-1}$ ($j \neq k$) into consideration to absorb common knowledge for further enhancing the learning. Thus, only finetuning the model $\mathbf{w}_c^k$ using $\mathcal{D}_s$ as the updated $\mathbf{w}_k^s$ is not a suitable solution. 
Second, federated learning needs to train each local model iteratively. Thus, it is difficult for the training of models to converge if the model architectures change frequently and dramatically. Furthermore, as each local model is a sub-structure of the original model $\mathbf{w}^{\prime}_s$, which may cause the lightweight model to miss the general knowledge that is learned by the original pretrained model. To avoid this issue, it requires that each model must keep its original architecture during the model fusion as well as aggregate general knowledge from the large model $\mathbf{w}_s^p$ stored at the server side.

Towards these three ends, we propose a novel \emph{structure-aware and knowledge-enhanced collaborative distillation} approach to learn a personalized aggregation model $\mathbf{w}_s^k$, which has the same network architecture as $\mathbf{w}_c^k$. 
In addition, even for the models uploaded from clients with different network architectures, the proposed structure-aware collaborative distillation approach still can boost the learning with the help from other local lightweight models and the pretrained model \emph{alternately} using the labeled data $\mathcal{D}_s$ stored on the server. 


\subsubsection{Structure-aware Similarity Learning}
In particular, when learning the model $\mathbf{w}_s^k$, we will take $\mathbf{w}_c^k$ as the leader, and $\{\mathbf{w}_c^j\}_{j=1}^{B-1}$ will be the helpers, as well as considering their network structures.
Intuitively, if the data distributions on two different clients are similar, the lightweight client models will also have similar network structures, which further generate similar outputs.
Thus, different helpers will contribute differently, and it is essential to distinguish the importance of helpers.
Based on this intuition, we propose a new approach to automatically assign a weight for each helper, and the weight is measured by the structure similarity between the leader and the helper.

Towards this target, we first obtain a binary mask vector $\boldsymbol{\theta}_k$ to indicate which weights are pruned by making a comparison between $\mathbf{w}_c^k$ and $\mathbf{w}_s^{\prime}$. Similarly, for each help, we can also obtain such a mask vector $\boldsymbol{\theta}_j$ ($j \neq k$). Then we can calculate a similarity score between $\boldsymbol{\theta}_k$ and $\boldsymbol{\theta}_j$, i.e.,
\begin{equation}
    \beta_j^k = \cos(\boldsymbol{\theta}_k, \boldsymbol{\theta}_j),
\end{equation}
where $\cos(\cdot, \cdot)$ denotes the cosine similarity. Since we have $B-1$ helpers, then we can obtain $B-1$ similarity scores. We finally normalize these scores and make $\sum_{j=1}^{B-1}\beta_j^k =1$.

\subsubsection{Knowledge-Enhanced Collaborative Distillation}
For training the designed structure-aware collaborative distillation method, we follow general knowledge distillation approaches~\cite{zhang2018deep} using the combination of the classification loss (i.e., cross entropy) and the Kullback–Leibler (KL) divergence of the weighted helpers and the pretrained model as follows:
\begin{equation}\label{eq:cd}
\begin{split}
    \mathcal{L}_s^k =& {\rm CE}\left(f_k(\mathbf{x}^s; \mathbf{w}_s^k), \mathbf{y}^s\right) + {\rm KL}(\boldsymbol{\alpha}^k_{avg}, \boldsymbol{\alpha}^k_s) \\
    &+ {\rm KL}(\boldsymbol{\alpha}_s^p, \boldsymbol{\alpha}^k_s),
\end{split}
\end{equation}
where $\boldsymbol{\alpha}^k_s$ is the soft target predictions or logits from the leader $\mathbf{w}_c^k$, $\boldsymbol{\alpha}_s^p$ represents the logits from the large pretrained model, and $\boldsymbol{\alpha}^k_{avg}$ represents the weighted average of logits from all helper models defined as $
\boldsymbol{\alpha}^k_{avg} = \sum_{j=1}^{B-1} \beta_j^k * \boldsymbol{\alpha}^j_{s}.
$

Note that $\mathbf{w}_s^k$ uses $\mathbf{w}_c^k$ as the initialization when optimizing Eq.~\eqref{eq:cd}. The proposed structure-aware collaborative distillation method can update all the models $\{\mathbf{w}_s^1, \cdots, \mathbf{w}_s^B\}$ in a parallel way if the computational capability of the server is large enough. Thus, such a design does not affect the computational efficiency too much for real-world applications.

After training each personalized model $\mathbf{w}_s^k$ on the server, these models will be then distributed to the corresponding local clients. Each local client will update and upload parameters again. This iterative procedure will continuously run until the models converge or achieve the maximum number of iterations. 

\section{Experiments}
In this section, we conduct a series of experiments on different types of datasets to validate the effectiveness of our proposed framework. We aim to answer the following research questions:
\begin{itemize}
    \item \textbf{RQ1}: Is the performance of our proposed framework comparable with or even better than that of state-of-the-art semi-supervised federated learning algorithms on different types of data?
    \item \textbf{RQ2}: With maintaining a comparable accuracy performance, how much communication cost does our framework save under the iterative communication mechanism?
    \item \textbf{RQ3}: How does each module take effects in the framework to achieve personalized efficiency federated learning under the semi-supervised setting?
\end{itemize}

\subsection{Experiment Setup}
To evaluate the generalization ability of our proposed framework, we validate it on two kinds of data. (1) \textbf{Image Datasets}. We conduct experiments for the image classification on {SVHN} and {CIFAR-10} in both IID and non-IID data distribution settings, respectively. 
(2) \textbf{Text Dataset}. We use {AG News} dataset to validate the performance of the text classification task.

This paper focuses on the semi-supervised federated learning setting, and thus, we choose {FedMatch}\cite{jeong2020federated}, {SSFL} \cite{zhang2020improving}, {FedMix} \cite{zhang2021semi}, {FedSEAL} \cite{bian2021fedseal}, and {SemiFL} \cite{diao2022semifl} as baselines.  

For image data, we use VGG19~\cite{simonyan2014very} as the pretrained model. For text data, we use BERT~\cite{devlin2018bert} as the large model. The lightweight models that we choose are a convolutional neural network (CNN) with six convolution layers and three linear layers (3,745,520
parameters) for the image data and a lightweight BERT model (71,766,890 parameters) for the text data. Note that \emph{we train both CNNs and lightweight BERT models from scratch}. 

\begin{table}[t]
\centering
\caption{Image classification performance comparing with semi-supervised federated learning baselines. }
\resizebox{0.95\columnwidth}{!}{
\begin{tabular}{l|cccc} 
\toprule 
Dataset&  \multicolumn{2}{c}{SVHN} & \multicolumn{2}{c}{CIFAR-10}\\
\hline
     Setting     & IID & non-IID & IID & non-IID\\
\hline
FedMatch \cite{jeong2020federated}   & 78.34\%   & 74.76\% & 64.70\%& 61.12\%\\
 SSFL \cite{zhang2020improving}  &76.06\%  & 70.29\% & 64.45\%      & 60.33\% \\
FedMix \cite{zhang2021semi}    & 78.45\%   & 71.76\% & 63.68\%       & 61.79\%\\

 FedSEAL \cite{bian2021fedseal}   & 72.64\% & 69.02\% &62.39\%      & 60.07\% \\
SemiFL \cite{diao2022semifl} & 84.65\%   & 82.15\%             & 70.79\%   &68.66\% \\

\midrule
\ours   & \textbf{85.31\%}    & \textbf{84.79\%} & \textbf{71.05\%}  & \textbf{69.81\%}  \\
\bottomrule 
\end{tabular}}
\label{tab:semi_result_image}
\end{table}

\subsection{Evaluation on Image Datasets}
\subsubsection{Performance Evaluation}
To answer \textbf{RQ1}, we choose the average local model text accuracy as our metric. We run each approach {three times} and report the \emph{average accuracy} in Table~\ref{tab:semi_result_image}.
For the image classification task, we have the following observations and analyses. 
Under the \textbf{IID setting}, the test accuracy of our work on both SVHN and CIFAR-10 datasets reaches the best accuracy at $85.31\%$ and $71.05\%$ respectively. However, FedMatch, FedMix, and SemiFL all leverage data augmentation techniques, which actually increase the number of training data and further benefit the performance improvement. Compared with baselines, {\ours} does not add any other synthetic data. Though with a shrunk model structure and fewer parameters, the collaborative distillation process enables each local model to extract general knowledge from other active local models.

Our proposed framework outperforms state-of-the-art semi-supervised federated learning baselines on SVHN and CIFAR-10 datasets under the \textbf{non-IID setting}. It implies that our personalized mechanism is able to address the data heterogeneity challenge better than other semi-supervised federated learning approaches because each local model is customized by the model structure compression and furthermore parameter fine-tuning with the local data.

\begin{figure}[!t]
        \centering
        \includegraphics[width=0.4\textwidth]{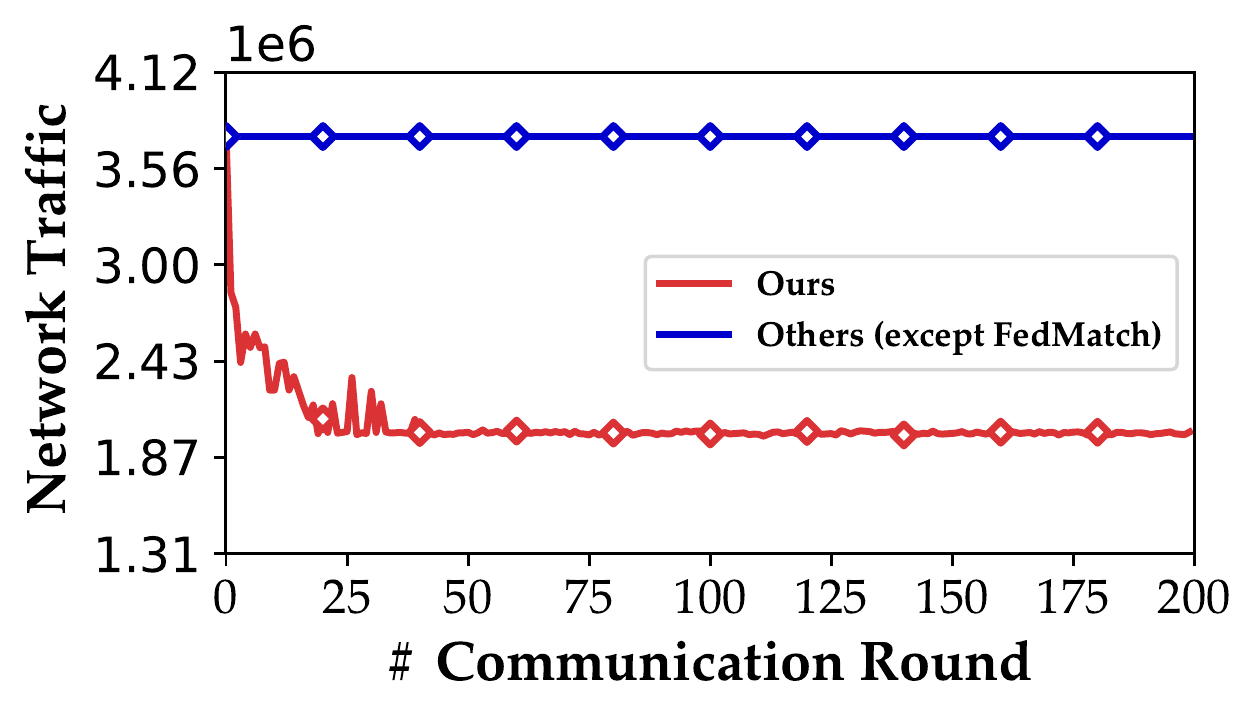}
        \caption{Communication cost analysis on CIFAR-10.}
        \label{fig:vgg}
\end{figure}

\subsubsection{Communication Efficiency Analysis}
To answer \textbf{RQ2}, we provide communication cost analysis on the CIFAR-10 dataset. The results can be found in Figure \ref{fig:vgg}. $x$-axis represents the number of communication rounds, and $y$-axis denotes the traffic between the server and active clients, i.e., the number of uploaded model parameters.

The blue lines are the network traffic of other baselines traffic except for FedMatch. Though there is a vector $\tau_t$ transmitted from the server to clients in FedSEAL~\cite{bian2021fedseal}, the value can be ignored compared with the model parameter size and total traffic. We report the communication cost saving ratio compared with the initial lightweight model by calculating (1 - our communication cost/conventional communication) $\times$ 100\%. \ours saves 44.9\% communication cost compared with baselines in average.

For \textbf{FedMatch}~\cite{jeong2020federated}, it reduces the communication cost by parameter decomposition and transmits the difference $\Delta \psi$ and $\Delta \sigma$. In their ``labels-at-server'' scenario, we get the average communication cost is 33.50\% when the active client ratio is 0.05 (F = 0.05). After fitting their setting by adjusting the active client ratio from 0.1 to 0.05, the average communication cost of {\ours} is only 22.45\%, which is 67.01\% of the communication cost of FedMatch.

\subsubsection{Ablation Study}

To answer \textbf{RQ3}, we conduct an ablation study to investigate the contributions of key modules -- knowledge enhancement, local compression, and collaborative distillation -- in our \ours framework on the image datasets. We use the following four reduced or modified models as baselines. The first three do not use pretrained models to enhance the learning, and the last one is to test the effectiveness of local client compression.

\textbf{AS-1}: Without using local model compression, pretrained large models, and collaborative distillation, the server aggregates local models using FedAvg and then fine-tunes the global model. AS-1 can be treated as \emph{Semi-FedAvg}. 

\textbf{AS-2}: Without using the designed collaborative distillation and pretrained large models, we just leverage a basic average approach via filling the pruned weights with 0, but keep the local model compression module. 

\textbf{AS-3}: Without using the pretrained large models to conduct knowledge enhancement learning, we only use lightweight models to conduct the system update with keeping all the other modules as \ours.

\textbf{AS-4}: Without using the local model compression, other operations are the same as our proposed approach. 

Table~\ref{tab:ablation} shows the experimental results. We can observe that reducing one or more key components in \ours makes the performance drop significantly. 
The results compared with AS-1 clearly show that the basic semi-supervised model aggregation is not enough for addressing the challenging semi-supervised federated learning problem.  
In AS-2, we apply client compression techniques to generate personalized local models but use a naive solution to aggregate the global model. Such a simple approach introduces extra errors in the model aggregation, which further leads to the worst performance among all baselines.
The performance of AS-3 is much better than that of AS-1 and AS-2. In AS-3, we use the designed collaborative distillation, which is an effective approach to obtain personalized client models. These results clearly confirm the superiority of the designed collaborative distillation.
AS-4 using pretrained models as prior knowledge outperforms the other three baselines, which proves that incorporating pretrained models is useful in federated learning. 
Besides, compared with \ours, the performance of AS-4 drops. This result demonstrates the necessity of client compression in the model design. 

\begin{table}[!t]
\centering
\caption{Ablation study on image datasets. }
\resizebox{0.45\textwidth}{!}{
\begin{tabular}{l|cccc} 
\toprule 
Dataset&  \multicolumn{2}{c}{SVHN} & \multicolumn{2}{c}{CIFAR-10}\\
\hline
     Setting     & IID & non-IID & IID & non-IID\\
\midrule
AS-1&  
72.50\% & 67.60\% & 
63.70\% &61.56\%  \\

AS-2 &
67.56\% & 65.04\%& 
59.74\% &59.05\% \\
AS-3 &
79.55\% & 77.52\%& 
63.67\% &63.47\%\\
AS-4 &
84.62\% & 80.16\%  &  
67.05\% & 65.71\%  \\
\midrule
\ours              &
\textbf{85.31\%}    &  \textbf{84.79\%} & \textbf{71.05\%}  & \textbf{69.81\%}  \\ 
\bottomrule 
\end{tabular}
}
\label{tab:ablation}
\end{table}

\subsection{Evaluation on Text Dataset}
To validate the generalization ability of the proposed \ours, we also conduct experiments on the text AG news dataset. 

\subsubsection{Performance Evaluation} 
\begin{figure}
\centering
\begin{subfigure}{\columnwidth}
  \centering
  \includegraphics[width=0.6\linewidth]{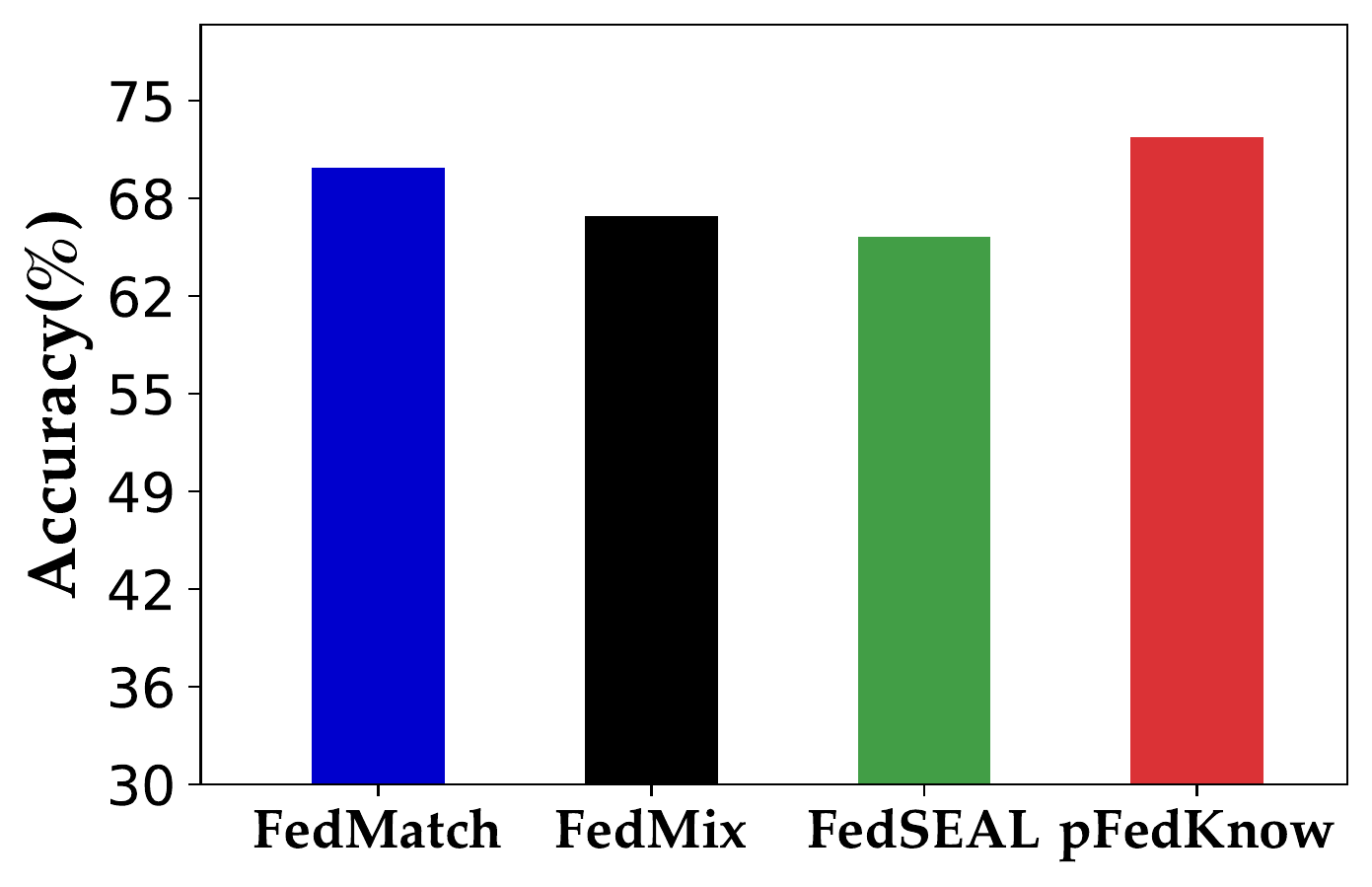}
  \caption{IID Setting}
  \label{fig:IID-Setting}
\end{subfigure}%

\begin{subfigure}{\columnwidth}
  \centering
  \includegraphics[width=0.6\linewidth]{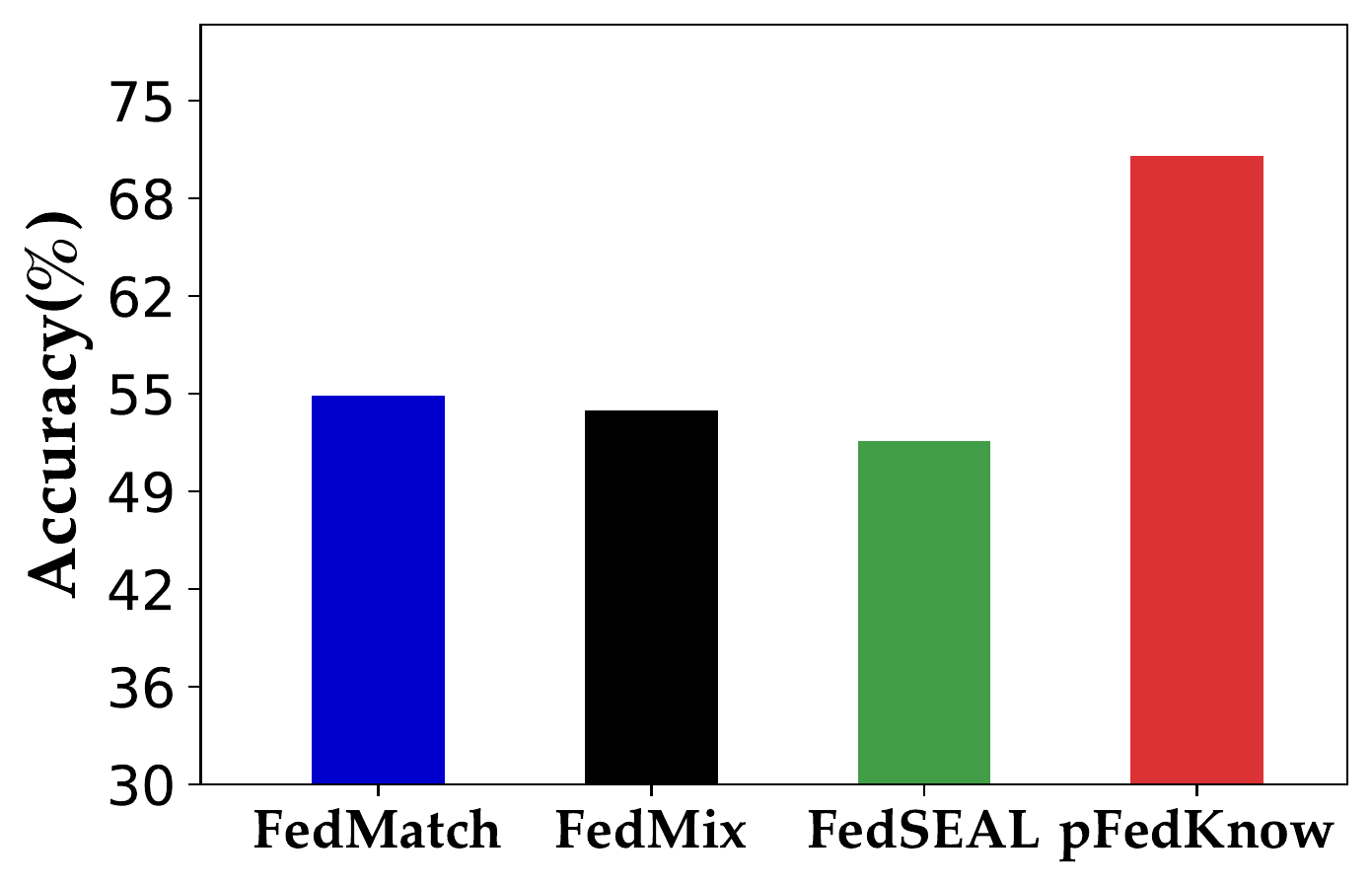}
  \caption{Non-IID Setting}
  \label{fig:sub2}
\end{subfigure}
\caption{Performance comparison on AG.}
\label{fig:text_compare}
\end{figure}

As there is limited work~\cite{lin2021fednlp} on federated learning in natural language processing, we choose the identical baselines with image tasks. 
However, according to the intuition and typical design for image tasks, SSFL~\cite{zhang2020improving} and FedMix~\cite{zhang2021semi} are not fit for text tasks. SSFL is based on replacing batch normalization with group normalization. In BERT-like models, layer normalization is leveraged rather than batch normalization. In FedMix, the operation on images does not make sense for the text tasks. Thus, we remove these two baselines to keep a fair comparison. Since baselines need augmented data as inputs, we follow the work~\cite{wei2019eda} to conduct data augmentation.
The results are shown in Figure~\ref{fig:text_compare}. We can observe similar results as the experiments on the image datasets. To highlight, \ours significantly increases the performance under the non-IID setting compared with the best baseline FedMatch. These results demonstrate the effectiveness of \ours again.

\begin{figure}[!t]

         \centering
         \includegraphics[width=0.4\textwidth]{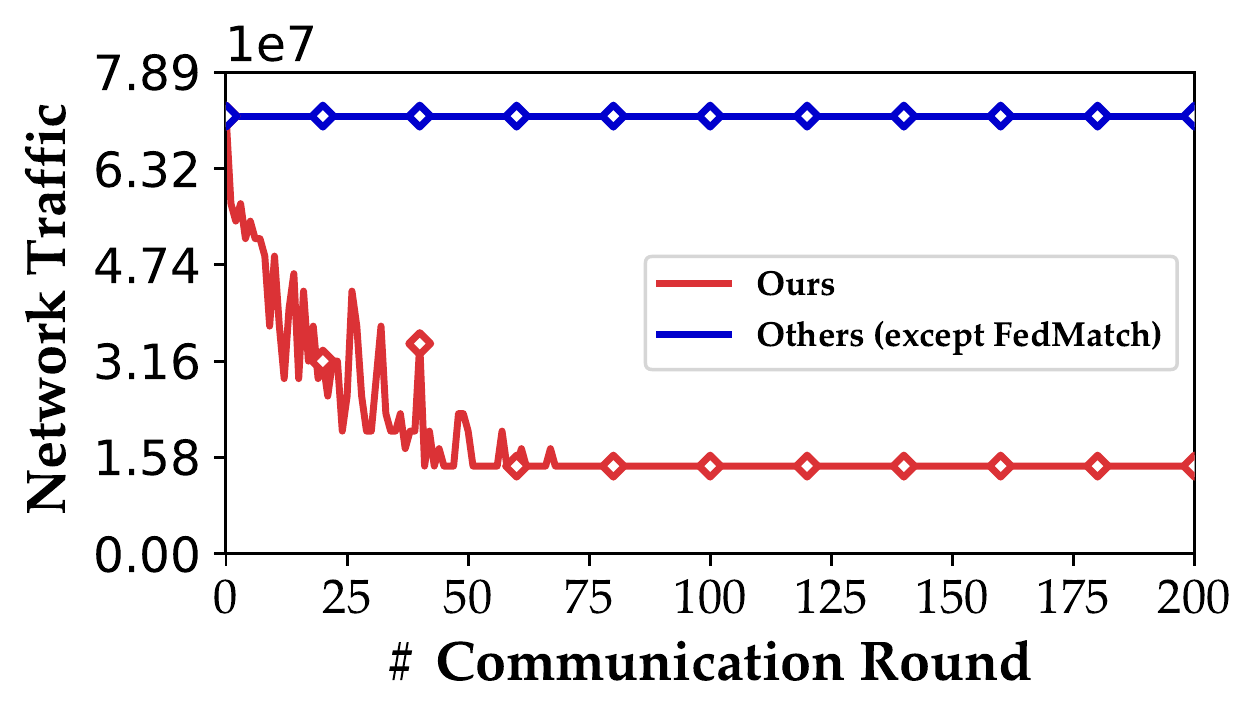}
         \caption{Communication cost analysis on AG.}
         \label{fig:bert}
\end{figure}

\subsubsection{Communication Efficiency Analysis}
We also analyze the communication cost on the text data. The results are shown in Figure~\ref{fig:bert}. In particular, the lightweight BERT model is compressed four times in total, which leads to the reduction of {73.6\%} communication cost in average.

\section{Related Work}

\subsection{Federated Learning in IoT}

Federated learning has been widely used in IoT. However, the limited bandwidth in the IoT environment poses a demanding requirement on the communication efficiency of FL. \cite{sattler2019robust} uses compression techniques such as sparsification and optimal Golomb encoding to reduce the traffic between clients and servers.  \cite{rothchild2020fetchsgd} uses a novel Count Sketch data structure to compress the uploaded gradient. CMFL~\cite{luping2019cmfl} reduces parameter exchange by avoiding uploading local updates which are not well-aligned with the global update. 
\cite{sun2020toward} tries to accelerate the model convergence and reduce the communication rounds by using gradient correction and batch normalization so that the optimizer can deal with the accumulated insignificant gradients properly.

The application of FL in IoT is also affected by data heterogeneity, which can be caused by sensor types, environments, tasks, etc. Some researchers have made efforts to mitigate this problem. For example, \cite{zeng2022heterogeneous} uses an efficient grouping method to assign heterogeneous clients to homogeneous groups and adapts a novel Sequential-to-Parallel training mode to coordinate the training within and among groups. \cite{wangtowards} uses data augmentation, hybrid training loss, and system-level training strategies to solve the data imbalance challenge. \cite{pang2020realizing} implements a reinforcement learning -based intelligent central server to capture the intricate patterns between clients with biased data and finally 
establishes an assignment plan with quasi-optimal performance. \cite{li2022data} down-samples similar users across factories to build homogeneous super nodes participating in FL training, which can effectively resist data heterogeneity.

\subsection{Semi-supervised Federated Learning}
Compared with traditional supervised federated learning~\cite{mcmahan2017communication, li2020federated}, semi-supervised federated learning is more practical and challenging. Several studies are proposed in the recent two years and focus on combining federated learning and semi-supervised learning techniques. 
In \cite{jeong2020federated}, a novel consistency loss is proposed to maximize the agreement between different local models, and disjoint learning is applied to extract reliable knowledge from the labeled and unlabeled data. \cite{zhang2020improving, zhang2021semi} design the group-based and weighted-based aggregation strategy to decrease the negative effect of data heterogeneity. In \cite{bian2021fedseal}, the authors add self-ensemble learning and complementary negative learning to federated learning. FedTriNet \cite{che2021fedtrinet} uses three networks and a dynamic quality control mechanism to generate high-quality pseudo labels for unlabeled data, and then the pseudo labeled and real labeled data are used together to retrain the model.  

Although these models can improve the prediction performance, none of them takes model personalization into consideration. Moreover, only a few models consider communication cost in the model design. This work aims to solve these challenging issues simultaneously and proposes an efficient personalized semi-supervised federate learning framework \ours.

\section{Conclusions}

In this paper, we propose a lightweight, effective, yet personalized semi-supervised federated learning framework named \ours, which can generally be applied on image and text classification tasks in IoT. We conduct experiments on both image and text datasets and demonstrate the effectiveness of \ours in dealing with data heterogeneity and communication cost challenges under the semi-supervised setting. 
\vspace{0.3in}

\bibliographystyle{plain}
\bibliography{ref}
\clearpage

\end{document}